\documentclass{article}
\makeatletter
\def\blindfootnote{\xdef\@thefnmark{}\@footnotetext}
\makeatother

\usepackage{microtype}
\usepackage{graphicx}
\usepackage{booktabs} 
\usepackage[numbers]{natbib}
\usepackage{svg}
\usepackage{wrapfig}

\usepackage{hyperref}

\usepackage[preprint]{neurips_2025}

\usepackage{amsmath}
\usepackage{amssymb}
\usepackage{mathtools}
\usepackage{amsthm}
\usepackage{xfrac}

\usepackage[capitalize,noabbrev]{cleveref}

\usepackage[textsize=tiny]{todonotes}

\usepackage{subcaption} 

\usepackage{enumitem} 

\usepackage{tabularx}

\usepackage{pifont}

\usepackage[utf8]{inputenc} 
\usepackage[T1]{fontenc}    
\usepackage{hyperref}       
\usepackage{url}            
\usepackage{booktabs}       
\usepackage{amsfonts}       
\usepackage{nicefrac}       
\usepackage{microtype}      
\usepackage{xcolor}         

\newtheorem*{remark}{Remark}

\title{Information-Theoretic Policy Pre-Training with Empowerment}

\author{
	Moritz Schneider\,\textsuperscript{\normalfont 1 2 3 *}
	\enskip
	Robert Krug\,\textsuperscript{\normalfont 1 2}
	\enskip
	Narunas Vaskevicius\,\textsuperscript{\normalfont 1 2}
	\\
	\textbf{Michael Volpp}\,\textsuperscript{\normalfont 1 2}
	\enskip
	\textbf{Luigi Palmieri}\,\textsuperscript{\normalfont 1 2}
	\enskip
	\textbf{Joschka Boedecker}\,\textsuperscript{\normalfont 3 4}\\
	\textsuperscript{\normalfont 1} Bosch Center for Artificial Intelligence \enskip
	\textsuperscript{\normalfont 2} Bosch Corporate Research \\
	\textsuperscript{\normalfont 3} University of Freiburg \enskip
	\textsuperscript{\normalfont 4} BrainLinks-BrainTools
}

\begin{document}

\maketitle

\begin{abstract}
	Empowerment, an information-theoretic measure of an agent's potential influence on its environment, has emerged as a powerful intrinsic motivation and exploration framework for reinforcement learning (RL).
Besides for unsupervised RL and skill learning algorithms, the specific use of empowerment as a pre-training signal has received limited attention in the literature.
We show that empowerment can be used as a pre-training signal for data-efficient downstream task adaptation.
For this we extend the traditional notion of empowerment by introducing discounted empowerment, which balances the agent's control over the environment across short- and long-term horizons.
Leveraging this formulation, we propose a novel pre-training paradigm that initializes policies to maximize discounted empowerment, enabling agents to acquire a robust understanding of environmental dynamics.
We analyze empowerment-based pre-training for various existing RL algorithms and empirically demonstrate its potential as a general-purpose initialization strategy: empowerment-maximizing policies with long horizons are data-efficient and effective, leading to improved adaptability in downstream tasks.
Our findings pave the way for future research to scale this framework to high-dimensional and complex tasks, further advancing the field of RL.

	\blindfootnote{
		\parbox{0.7\textwidth}{
			\textsuperscript{*}Correspondence to: \texttt{moritz.schneider@de.bosch.com}
		}
	}
\end{abstract}

\section{Introduction} \label{sec:introduction}
Pre-training has emerged as a critical strategy in reinforcement learning (RL) to improve sample efficiency, stability, and generalization, particularly in environments where reward signals are sparse or delayed \cite{levine_offline_2020,xie_pretraining_2022}. By enabling agents to acquire useful priors about the structure of the environment or about transferable behavioral patterns, pre-training can significantly increase data efficiency during downstream task adaptation \cite{xie_pretraining_2022,springenberg_offline_2024}. Empowerment \cite{klyubin_all_2005,klyubin_empowerment_2005,salge_empowermentintroduction_2014}, an information-theoretic measure of an agent's potential influence over future state distributions, offers a principled approach to unsupervised pre-training \cite{mohamed_variational_2015,karl_unsupervised_2019,leibfried_unified_2019,zhao_efficient_2020}. Unlike tasks with externally given rewards, empowerment does not rely on external rewards but instead encourages the agent to explore and occupy states from which it retains a high degree of control in the environment. This intrinsic motivation enables the development of broadly applicable behaviors that can be rapidly adapted to a variety of downstream objectives.

Unsupervised pre-training has become a cornerstone in the development of foundation models in both computer vision \cite{radford_learning_2021,girdhar_imagebind_2023,oquab_dinov2_2024} and natural language processing \cite{brown_language_2020,gemini_team_gemini_2024,gemini_team_gemini_2024-1}, where large-scale models are trained on vast amounts of unlabeled data to learn general-purpose representations. Despite these advances, a comparable paradigm has not yet been established within RL. One promising domain for RL pre-training is video data \cite{yang_position_2024,mccarthy_towards_2024}, which naturally embodies sequential dynamics and agent-environment interactions. Utilizing such data would be additionally useful due to the difficulty of collecting real-world data for RL agents as video data is broadly available. However, the absence of explicit reward signals in such data precludes the direct application of traditional RL techniques. This limitation necessitates the use of unsupervised RL approaches, where intrinsic objectives—such as curiosity~\cite{pathak_curiosity-driven_2017}, information gain~\cite{sekar_planning_2020,rhinehart_information_2021} or empowerment~\cite{mohamed_variational_2015,karl_unsupervised_2019}—serve as substitutes for task-specific rewards. These unsupervised signals enable agents to learn useful representations and behaviors that can generalize across tasks, potentially paving the way for foundation models in RL analogous to those in vision and language. We show that empowerment can be effectively employed as a pre-training signal, enabling agents to learn useful representations and behaviors that can be rapidly adapted to various downstream tasks.

A key distinction between our work and prior research on empowerment lies in the role of empowerment-based objectives. Previous studies have predominantly employed empowerment and mutual information maximization as intrinsic reward signals aimed at enhancing exploration during RL \cite{houthooft_vime_2016,leibfried_unified_2019,becker-ehmck_exploration_2021,zhang_metacure_2021} or complete unsupervised control \cite{mohamed_variational_2015,karl_unsupervised_2019,zhao_efficient_2020}, in contrast to targeting improved initialization of policies before actual downstream task learning. Its application as a standalone pre-training signal remains largely underexplored. In other cases, empowerment has been leveraged to facilitate the unsupervised acquisition of diverse skills, which are typically pre-trained and subsequently utilized in downstream tasks \cite{sharma_dynamics-aware_2020,laskin_contrastive_nodate,eysenbach_diversity_2019}. However, \citet{eysenbach_information_2022} demonstrated that, contrary to conventional assumptions, maximizing mutual information may not necessarily result in the development of distinct skills, but rather serves to generate a beneficial initialization for skill acquisition. We argue that such an initialization would be beneficial for a general policy itself which is not dependent on a skill identifier.

In short, we address these gaps by the following key contributions:
\begin{itemize}
    \item We propose a generic framework for unsupervised pre-training in RL based on empowerment, emphasizing its utility for initializing policies in downstream tasks.
    \item We introduce discounted empowerment as a flexible objective that balances short- and long-term control, enabling a simple pre-training strategy without the need to tune a specific horizon for multi-step empowerment.
    \item We demonstrate through experiments that empowerment-based pre-training with our discounted empowerment reward accelerates the adaption process and improves learning efficiency on a variety of downstream RL tasks.
\end{itemize}
\section{Related Work} \label{sec:related_work}
\textbf{Unsupervised Control.}\label{para:empowerment}
Recent research in intrinsically motivated RL has focused on leveraging information-theoretic principles to drive agent behavior without extrinsic rewards. \citet{klyubin_all_2005,klyubin_empowerment_2005} introduced empowerment as a universal, agent-centric measure of control, quantifying an agent's potential to influence its future states via channel capacity between actions and sensor states. This concept was further operationalized in subsequent works \cite{klyubin_keep_2008,salge_approximation_2013,salge_changing_2014,salge_empowerment_2017}, where empowerment served as a general-purpose utility function for sensorimotor systems, promoting behaviors that maximize an agent's action-perception possibilities. Building on this foundation, \citet{karl_unsupervised_2019} and \citet{mohamed_variational_2015} proposed variational approaches towards empowerment maximization. Both methods enable real-time control via a tractable, unsupervised variational approximation of empowerment. \citet{houthooft_vime_2016}, \citet{leibfried_unified_2019} and \citet{navneet_madhu_kumar_empowerment-driven_2018} introduced empowerment-based exploration strategies, which leverage empowerment to guide exploration in RL tasks additionally to the extrinsically given reward signal. While empowerment can clearly excel as an exploration signal, such a use case does not lead to an intrinsically competent agent as the ultimate goal is still to maximize the extrinsic reward.
\citet{liu_behavior_2021} introduced APT, an entropy maximization method leveraging contrastive representation learning for efficient pre-training and adaptation in visual RL tasks.
\citet{mendonca_discovering_2021} presents LEXA, a model-based approach that trains separate explorer and achiever policies on imagined trajectories using a learned world model, exploring diverse states and enabling goal-conditioned behavior via learned latent distances.
\citet{yarats_reinforcement_2021} introduced Proto-RL, which utilizes prototypical representations and particle-based entropy maximization for effective exploration and representation learning in image-based RL.
In another direction, \citet{myers_learning_2024} proposed assistance policies that, instead of maximizing the empowerment of the agent itself, learn to preserve human autonomy through human empowerment maximization. Even though, all those works focus on unsupervised and empowerment-based control, none of these actually use empowerment-based policies to initialize a policy for downstream tasks. Instead, they focus on completely independent unsupervised task-solving or exploration.

Similarly, \textbf{Unsupervised Skill Discovery}\label{para:mi_skill} algorithms focus on learning a repertoire of diverse behaviors without explicit external rewards by including learned skill identifiers. Methods like VALOR \cite{achiam_variational_2018}, VIC \cite{gregor_variational_2017}, DIAYN \cite{eysenbach_diversity_2019} and DADS \cite{sharma_dynamics-aware_2020} leverage variational lower bounds on mutual information to learn distinguishable skills.
Contrastive Intrinsic Control (CIC) \citep{laskin_contrastive_nodate} employs contrastive learning between state transitions and skills to learn behavior embeddings, using their entropy as an intrinsic reward.
\citet{liu_aps_2021} (who builds upon APT) and \citet{hansen_fast_2020} incorporate successor features while \citet{park_controllability-aware_2023} moves away from mutual information and instead optimizes a certain Wasserstein distance to learn skills.
In a similar approach, \citet{zheng_can_2025} connects successor features, contrastive learning and mutual information maximization.
\citet{eysenbach_information_2022} examined the relationship between skill learning and the geometry of state marginal distributions. The paper reveals that skill learning with mutual information maximization leads to favorable initializations of skill policies and the learned skills may not cover all possible behaviors.
Choreographer \cite{mazzaglia_choreographer_2022}, decouples exploration from skill discovery and uses methods like VQ-VAE\cite{van_den_oord_neural_2017} to learn a codebook of skills based on world model states.
In contrast to those skill discovery approaches, our work does not focus on learning a set of individual skill policies, but rather on learning a single policy that is capable of quickly adapting to any downstream task. Our goal is to learn a good initialization for a single policy that can be fine-tuned fast on any downstream task.

\textbf{Meta-Reinforcement Learning}\label{para:metarl} is a subfield of RL that focuses on training agents to learn new tasks quickly by leveraging prior experience. The goal is to enable agents to adapt to new environments or tasks with minimal data and time. \citet{finn_model-agnostic_2017} proposed Model-Agnostic Meta-Learning (MAML), which trains a model on a variety of tasks such that it can quickly adapt to new tasks with only a few gradient updates. It does so without introducing additional learnable parameters and purely with gradient descent.
\citet{wang_learning_2016} and \citet{duan_rl2_2016} are model-agnostic as both methods rely on recurrent neural networks to learn a policy that can adapt to new tasks quickly.
\citet{zintgraf_varibad_2020} introduces variBAD, a method to approximate Bayes-optimal behavior, which uses meta-learning to utilize knowledge obtained in related tasks and perform approximate inference in unknown environments.
The only work known to us which combines empowerment with meta learning is \citet{zhang_metacure_2021} who combine meta RL with empowerment-based exploration.
All of these methods can be interpreted as initializing a policy for efficient adaptation to unseen tasks. In other words, all learn an initialization of the policy that is closer to the optimal policy for the new task than a random initialization. Our work is related in the sense that we utilize empowerment to learn similar initializations. However, in contrast to meta-RL, an empowerment-based initialization is not based on extrinsically-defined pre-training tasks and it does not require a distribution of tasks for pre-training. Instead, it yields a generic initialization that can be used for any downstream task in a given MDP.

\section{Policy Pre-Training with Empowerment} \label{sec:optimal_init}
In this section, we introduce our approach for empowerment-based pre-training of RL policies. The goal is to initialize the policy in a way that enables quick adaptation to every downstream task at hand.

\subsection{Reinforcement Learning} \label{sec:rl_intro}
Reinforcement learning (RL) is a framework for decision-making in which an agent learns optimal behavior by interacting with an environment and receiving feedback in the form of rewards. The environment is typically modeled as a Markov Decision Process (MDP) defined by the tuple~$(\mathcal{S}, \mathcal{A}, \mathcal{P}, \rho_0, R, \gamma)$, with the set of states~$\mathcal{S}$ and the set of actions~$\mathcal{A}$. $\mathcal{P}(\mathbf{s'} \mid \mathbf{s}, \mathbf{a})$ denotes the transition probability density, which defines the probability of transitioning to state~$\mathbf{s}' \in \mathcal{S}$ when choosing action~$\mathbf{a} \in \mathcal{A}$ in state~$\mathbf{s} \in \mathcal{S}$ where $\mathcal{P}: \mathcal{S} \times \mathcal{A} \rightarrow \Delta (\mathcal{S})$. Here, $\Delta (\mathcal{S})$ is the space of probability distributions over $\mathcal{S}$. $R(\mathbf{s})$ is a scalar function which specifies the immediate reward received when reaching state~$\mathbf{s}$. The goal of an agent is to find a policy~$\pi(\mathbf{a} \mid \mathbf{s})$, defining a conditional probability distribution over actions given a state, which maximizes the expected future cumulative discounted reward for a given state $\mathbf{s}$:
\begin{equation}
    J(\pi) = \mathbb{E} \left [ \sum_{t=0}^\infty \gamma^t R(\mathbf{s}_{t+1}) \right ].
\end{equation}
Here, $\mathbf{s}_t \in \mathcal{S}$ denotes the state at time~$t$ and the expectation is w.r.t.~the state distribution induced by the policy $\pi$ and the transition probability density $\mathcal P$, starting from an initial state distribution $\rho_0(\mathbf{s}_0)$. The discount factor~$\gamma \in [0, 1)$ weights future rewards relative to immediate ones.

\subsection{Empowerment} \label{sec:empowerment}
We start by defining the information-theoretic concept of channel capacity. Intuitively, channel capacity quantifies the maximum number of bits of information that can be reliably transmitted over a communication channel \citep{cover_elements_2005}. Mathematically, a channel with input $X$ and output $Y$ is characterized by the conditional probability density $p(y \mid x)$, and we define its channel capacity~$\mathcal{C}$ as the maximum mutual information between output and input,
\begin{equation}
    \mathcal{C} \doteq \max_{p(x)} \mathcal{I}(X; Y) = \max_{p(x)} \left ( \mathcal{H}(X) - \mathcal{H}(X \mid Y) \right ),
\end{equation}
where the maximum is taken over all possible input distributions $p(x)$ and $\mathcal{H}(X)$ denotes the entropy of $X$.

The empowerment formalism applies this idea to sensorimotor systems. In the context of our RL setup, the channel is represented by the transition dynamics~$\mathcal{P}(\mathbf{s}' \mid \mathbf{s}, \mathbf{a})$, considering the action~$\mathbf{a}$ as the input and the next state~$\mathbf{s'}$ as the output. The contextual empowerment~$\mathcal{E}$ for a state~$\mathbf{s}$ is then defined as the channel capacity between next states and actions given the current state~$\mathbf{s}$ as context \cite{klyubin_keep_2008,salge_empowermentintroduction_2014}:
\begin{equation}
    \label{eq:empowerment}
    \mathcal{E}(\mathbf{s}) \doteq \max_{p(\mathbf{a})} \mathcal{I}(\mathbf{s}'; \mathbf{a} \mid \mathbf{s}) \doteq \max_{p(\mathbf{a})} \mathbb{E}_{p(\mathbf{a}) \mathcal{P}(\mathbf{s}' | \mathbf{s}, \mathbf{a})} \left[\log{\frac{\mathcal{P}(\mathbf{s}'\mid \mathbf{s}, \mathbf{a})}{\mathcal{P}(\mathbf{s}' \mid \mathbf{s})}}\right].
\end{equation}
Since empowerment is computed over all possible policies~$p(\mathbf{a})$, it becomes a property of the state~$\mathbf{s}$ alone, independent of any specific agent behavior or policy. Consequently, empowerment characterizes the environment's intrinsic structure, capturing the degree of influence an agent can potentially exert from a given state, irrespective of its current policy.

Empowerment can be interpreted as a proxy for an agent's preparedness \cite{salge_empowermentintroduction_2014}, reflecting its ability to influence future states in the environment. From this perspective, a highly empowered agent is one that possesses a broad repertoire of future options, enabling it to navigate toward states that are potentially advantageous or preferable, even in the absence of explicit external rewards. This capability implies a form of competence, whereby the agent acquires an implicit understanding of the environmental structure. By prioritizing states that maximize the agent's future influence, the agent demonstrates a form of intrinsic preparedness.

We assume that an agent has maximal competence if it has figured out a policy that achieves or maximizes the empowerment available in its environment. The agent's empowerment can be increased by either (1) going to more empowered states or by (2) changing the agent or environment \cite{salge_empowermentintroduction_2014}. In the standard RL setup, it is not possible to change the sensors and available actions of the agent, nor it is possible to change the environment beyond the state-space of the MDP. Thus, the only way to increase empowerment is by going to more empowered states. Thus, it would be natural to pre-train agents to maximize empowerment in the environment (which is depicted on the left part of Figure~\ref{fig:method}).

\paragraph{$n$-step Empowerment.} The concept of $n$-step empowerment~$\mathcal{E}_n$ generalizes the traditional notion of empowerment by extending the temporal scope from immediate state-action transitions to sequences of transitions over a fixed horizon of $n$ steps. Specifically, rather than measuring the mutual information between a single action~$\mathbf{a}_t$ and the resulting next state~$\mathbf{s}_{t+1}$, $n$-step empowerment evaluates the mutual information between a sequence of actions~$\mathbf{a}_{t:t+n} \in \mathcal{A}^n$ and the resulting state~$\mathbf{s}_{t+n+1}$, conditioned on the initial state~$\mathbf{s}_t$:
\begin{equation}
    \mathcal{E}_{n} (\mathbf{s}_t) \doteq \max_{p(\mathbf{a}_t, \dots, \mathbf{a}_{t+n})} \mathcal{I}(\mathbf{s}_{t+n+1}; \mathbf{a}_{t:t+n}| \mathbf{s}_t).
\end{equation}
This formulation enables the agent to assess its potential influence over the environment across a longer time horizon, thereby capturing more temporally extended notions of controllability. By incorporating multi-step dynamics, $n$-step empowerment provides a richer, more comprehensive measure of how much control an agent can exert over its future, beyond the scope of immediate transitions.

\subsection{Discounted Empowerment} \label{sec:discounted_empowerment}
\begin{figure*}[t]
	\centering
	\hfill
	\begin{subfigure}[t]{0.16\linewidth}
		\includegraphics[width=\linewidth]{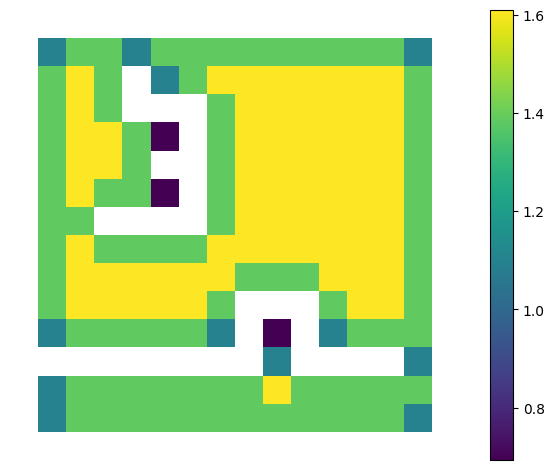}
	\end{subfigure}
	\hfill
	\begin{subfigure}[t]{0.16\linewidth}
		\includegraphics[width=\linewidth]{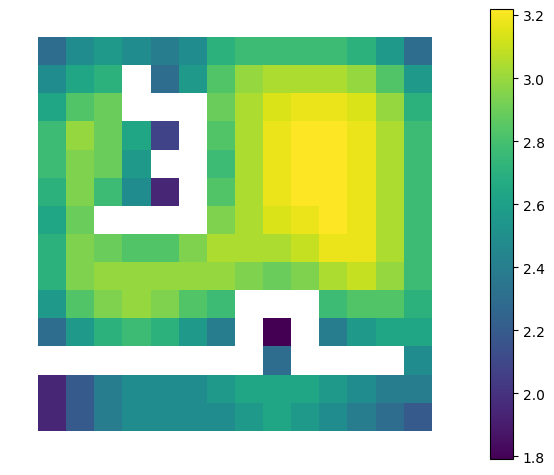}
	\end{subfigure}
	\hfill
	\begin{subfigure}[t]{0.16\linewidth}
		\includegraphics[width=\linewidth]{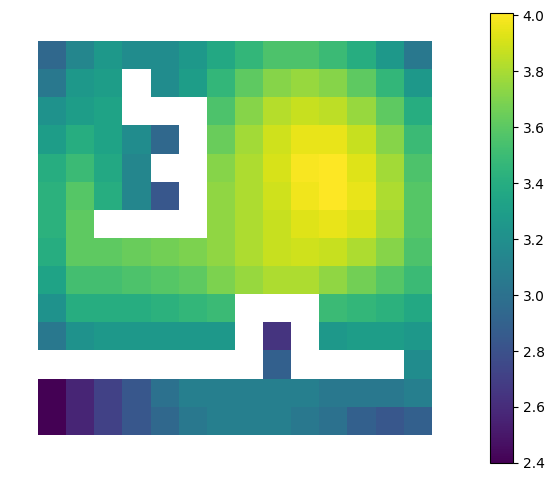}
	\end{subfigure}
	\hfill
	\begin{subfigure}[t]{0.16\linewidth}
		\includegraphics[width=\linewidth]{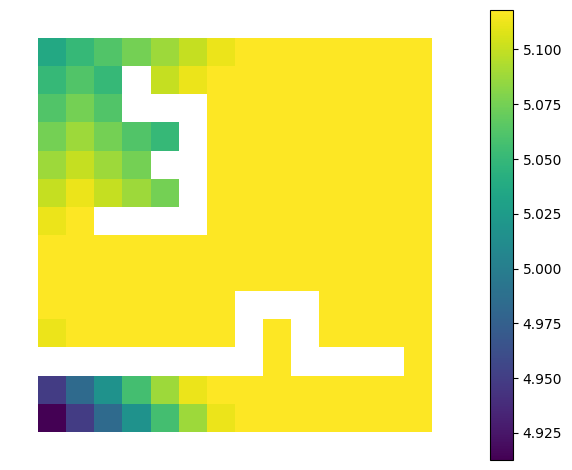}
	\end{subfigure}
	\hfill
	\begin{subfigure}[t]{0.16\linewidth}
		\includegraphics[width=\linewidth]{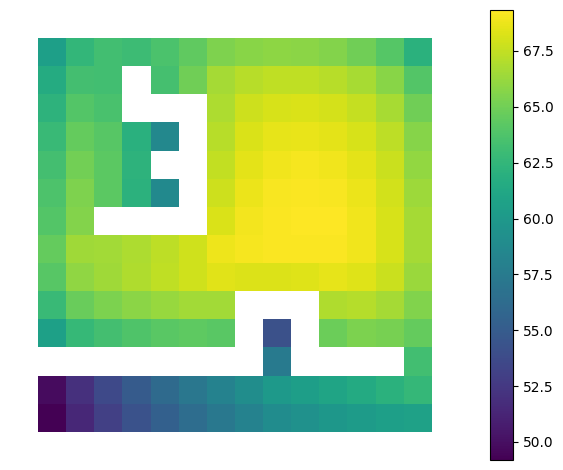}
	\end{subfigure}
	\hfill
	\begin{subfigure}[t]{0.16\linewidth}
		\includegraphics[width=\linewidth]{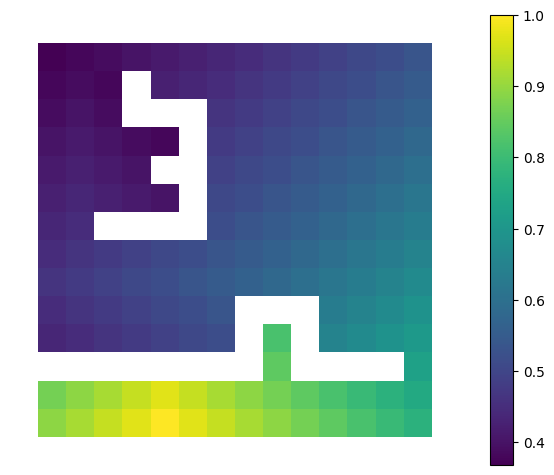}
	\end{subfigure}
	\hfill
	\caption{\label{fig:gridworld} Empowerment-values of our deterministic gridworld environment. The empowerment values are calculated for $1$-, $3$-, $5$ and $32$-steps and for the discounted case (from left to right). The $32$-steps empowerment grid shows a mostly uniform empowerment landscape due to the issue that most other states can be reached from any state in a horizon of $32$ steps. The last image shows the reward map for an exemplary goal state of the environment.}
	\vspace{-2ex}
\end{figure*}

A fundamental limitation of $n$-step empowerment with large horizon values arises from the agent's ability to access an increasingly uniform set of reachable states. As $n$ grows, the empowerment definition may include the possibility to reach virtually any state from any starting point, resulting in a uniformly flat empowerment landscape \cite{salge_empowermentintroduction_2014}. In such cases, empowerment degenerates into a trivial signal thereby losing its ability to differentiate states based on their available future options. Such uniform empowerment landscape for a deterministic gridworld can be seen in the fourth grid of Figure~\ref{fig:gridworld}.

To address this issue, we introduce \textit{discounted empowerment}: Instead of training the agent on a specific $n$-step empowerment horizon, we evaluate agents on discounted versions of multiple $n$-step horizons. Similar to $n$-step empowerment, it is a measure of the amount of control an agent has over its environment, taking into account different horizons with the discount factor~$\lambda \in (0, 1]$ and total length~$H$. It is defined as the discounted sum of channel capacities between the agent's actions and the future states of the environment, with the discount factor~$\lambda$ applied to empowerment values of larger horizons. The discounted empowerment can be expressed as:
\begin{equation}
    \mathcal{E}_{\lambda} (\mathbf{s}_i) \doteq \sum_{k=0}^{H} \lambda^{k}  \max_{p(\mathbf{a}_i, \dots, \mathbf{a}_{i+k})} \mathcal{I}(\mathbf{s}_{i+k+1}; \mathbf{a}_{i:i+k}| \mathbf{s}_i).
\end{equation}
Here, $H$ denotes the episode length. This formulation allows us to capture the short- and long-term effects of actions on the state distribution, which we assume are both crucial for effective pre-training. Consequently, the concept of discounted empowerment obviates the necessity of determining an appropriate empowerment horizon tailored to the specific characteristics of the environment.
It is important to note that while our discount factor $\lambda$ resembles the RL discount factor $\gamma$, the two serve distinct purposes and are not equivalent.

\subsection{Pre-Training with Empowerment}
\begin{figure*}[t]
	\centering
	\begin{subfigure}[t]{\linewidth}
		\includegraphics[width=\linewidth]{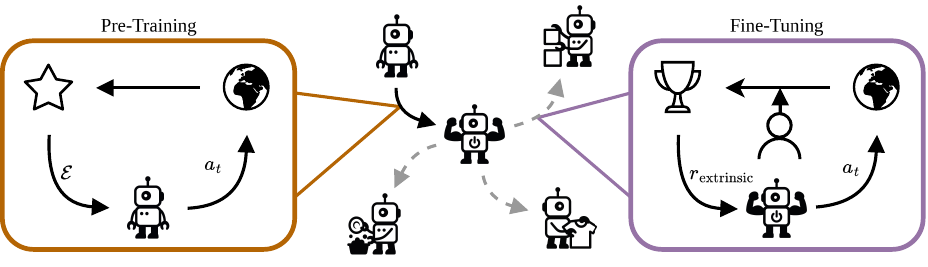}
	\end{subfigure}
	\caption{\label{fig:method}
	\textbf{Empowerment-based Pre-Training.} The untrained agent is pre-trained to optimize empowerment $\mathcal{E}$ in an environment-centric manner, which allows it to learn a policy that can be fine-tuned for specific tasks. The initialization achieved by this pre-training is expected to be closer to the optimal policy than a random initialization, as it has already learned to achieve options that are helpful for downstream tasks. The empowerment value is solely based on the environment characteristics without further human input whereas the extrinsic fine-tuning reward is based on expert human knowledge of the task.}
	\vspace{-2ex}
\end{figure*}
To pre-train an agent, we first compute the empowerment values for each state within the environment, quantifying the agent's potential influence over future states without depending on a specific policy. These empowerment values are then employed as intrinsic rewards $R(\mathbf{s}_t) = \mathcal{E}_i(\mathbf{s}_t)$, enabling the agent to be pre-trained using a RL algorithm of choice with the objective of maximizing empowerment.
Here, $\mathcal{E}_i$ can be one-step empowerment~$\mathcal{E}$, $n$-step empowerment~$\mathcal{E}_n$ or discounted empowerment~$\mathcal{E}_\lambda$.
The resulting pre-trained policy~$\pi_\mathcal{E}$, optimized for empowerment maximization, is subsequently deployed directly to a downstream task without requiring additional modifications to the policy. Notably, the MDP of the downstream task differs solely in its extrinsic reward function, which is constructed using expert human knowledge to align with the specific objectives of the task. This scheme is illustrated in Figure~\ref{fig:method}.
\section{Experiments} \label{sec:experiments}
We perform experiments in a discrete gridworld nevigation environment with $5$ discrete actions for moving forward, backward, left, right and waiting. The agent aims to reach a certain goal state.
This environment provides a controlled and interpretable setting where empowerment can be computed efficiently.
From Equation~\ref{eq:empowerment} it is easy to see, that in the deterministic case, empowerment of a state $\mathbf{s}_t$ is determined by the logarithm of the number of states reachable within a specified horizon, offering a straightforward and tractable measure of an agent's influence over its environment.
For the stochastic case, we employ the Blahut-Arimoto algorithm \cite{blahut_computation_1972, arimoto_algorithm_1972}, a well-established method for calculating channel capacity in probabilistic systems. Stochasticity is introduced by allowing the agent to move to the left or to the right relative to its current direction with a predefined probability, thereby incorporating an element of uncertainty into the agent's actions.

In our experiments, we compare five distinct approaches: on-policy REINFORCE~\cite{williams_simple_1992} agents with and without a baseline, on-policy Actor-Critic (AC) agents incorporating a critic trained with temporal-difference (TD) learning~\cite{sutton_reinforcement_2018}, on-policy Proximal Policy Optimization (PPO)~\cite{schulman_proximal_2017} agents incorporating generalized advantage estimation~\cite{schulman_high-dimensional_2016}, and off-policy Deep Q-Network (DQN)~\cite{mnih_human-level_2015} agents. PPO and DQN agents are trained using raw RGB image representations of the environment, whereas REINFORCE and AC agents operate on a low-dimensional state representation based on a one-hot encoding of the agent's location. For the REINFORCE and AC agents, each pre-trained model is fine-tuned individually on every possible goal state within the environment, resulting in a total of $835$ fine-tuning runs for each algorithm. In contrast, the RGB observations used by PPO and DQN agents include both the agent's state and the goal position, enabling fine-tuning in an episodic manner with goals that are randomly sampled from the set of all valid states. This allows us to evaluate the benefit of empowerment within two distinct tasks: (1) episodic fine-tuning with randomly sampled goals on RGB images (PPO and DQN), and (2) individual fine-tuning on every possible goal state within the environment on the encoded agent position (REINFORCE and AC). For more details on our experimental setup we refer to Appendix~\ref{sec:appx_implementation_details}.

\subsection{Capacity-Achieving vs. Capacity-Maximizing Policies}
The distinction between capacity-achieving and capacity-maximizing policies lies in their objectives and outcomes.
A capacity-achieving policy~$\pi^*(\mathbf{a} | \mathbf{s})$ (often referred to as source policy \cite{mohamed_variational_2015,karl_unsupervised_2019}) is one that is needed to calculate the empowerment value~$\mathcal{E}(\mathbf{s})$ for a given state~$\mathbf{s}$ (i.e. the policy $\pi$ which achieves empowerment in Equation~\ref{eq:empowerment}).
It is the outcome of the Blahut-Arimoto algorithm \cite{blahut_computation_1972,arimoto_algorithm_1972} and the optimal one from previously reported results on optimal initializations for unsupervised skill learning methods \cite{eysenbach_information_2022}, cf.~Appendix \ref{sec:appx_info_geom_empowerment}.
Such a policy visits each state equally often under the transition dynamics~$\mathcal{P}(\mathbf{s}' | \mathbf{s}, \mathbf{a})$. 
It does not seek states with high empowerment values but rather focuses on optimizing the mutual information~$\mathcal{I}(\mathbf{s}'; \mathbf{a} | \mathbf{s})$ at that specific state. 
On the other hand, a capacity-maximizing policy refers to a policy that actively seeks to maximize empowerment over time by navigating the environment to reach states with higher empowerment values (i.e. directly maximizing the given empowerment values). Previous work advocate for capacity-maximizing policies as the optimal ones for empowerment-based, unsupervised control \cite{klyubin_all_2005,klyubin_empowerment_2005,klyubin_keep_2008,salge_empowermentintroduction_2014}.

To assess which pre-training approach is more effective for fine-tuning on downstream task, we empirically compare the sample efficiency during fine-tuning of agents pre-trained with capacity-maximizing policies to those pre-trained with capacity-achieving policies on the gridworld navigation task. All agents are trained using the REINFORCE algorithm \cite{williams_simple_1992}. The left panel of Figure~\ref{fig:exp_horizon_and_capacity} shows that the capacity-maximizing policies are more data-efficient than the capacity-achieving policies. Nevertheless, both methods yield pre-trained agents that outperform the baseline when fine-tuned to downstream tasks.
In the following, we will focus on capacity-maximizing policies.

\subsection{Influence of Empowerment Horizon}
\begin{figure*}[t]
	\centering
	\begin{subfigure}[t]{\linewidth}
		\includegraphics[width=\linewidth]{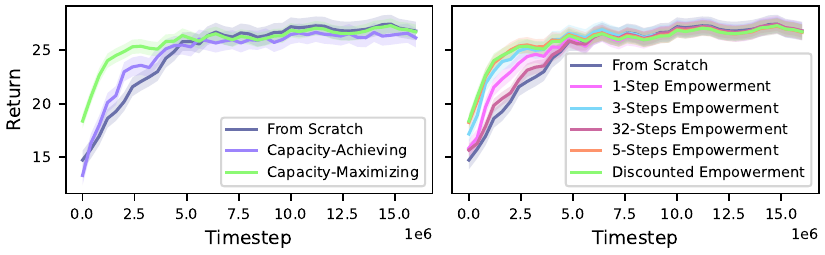}
	\end{subfigure}
	\caption{\label{fig:exp_horizon_and_capacity} REINFORCE training curves for fine-tuning on individual goal states in a deterministic grid-world environment. \textbf{Left panel}: Both capacity-achieving and capacity-maximizing policies outperform the baseline in terms of data efficiency, with the capacity-maximizing agent performing best. \textbf{Right panel}: Comparison of the performance of agents pre-trained with $n$-step empowerment with different empowerment horizons and using our proposed discounted empowerment reward. Discounted empowerment performs favorably, without the need to tune the horizon length.}
	\vspace{-2ex}
\end{figure*}
Next, we evaluate our proposed formalism of discounted empowerment by investigating the impact of the empowerment horizon on the downstream performance of REINFORCE agents.
Specifically, the agents are pre-trained using varying empowerment horizons and subsequently fine-tuned across all possible goal states within the environment.
The aggregated results, as illustrated in the right panel of Figure~\ref{fig:exp_horizon_and_capacity}, again demonstrate that agents pre-trained with empowerment consistently achieve superior data-efficiency during fine-tuning. These findings suggest that extended (effective) horizons $n > 1$ enable more effective exploration.
Moreover, we observe that the benefits of pre-training diminish as the empowerment horizon shortens. Conversely, if the horizon is too long, performance also degrades. This suggests the existence of an optimal $n$-step empowerment horizon that is environment-dependent.
Our proposed formulation instead uses a discounted empowerment objective, which balances short- and long-term empowerment through a discount factor~$\lambda$. Notably, we found that performance is sensitive to the choice of $n$, but largely insensitive to $\lambda$. This allows us to fix $\lambda = 0.95$ across all experiments, significantly reducing the burden of hyperparameter tuning. Thus, discounted empowerment is the most effective choice for pre-training.

\subsection{Deterministic Gridworld}
\begin{figure*}[t]
	\centering
	\begin{subfigure}[t]{\linewidth}
		\includegraphics[width=\linewidth]{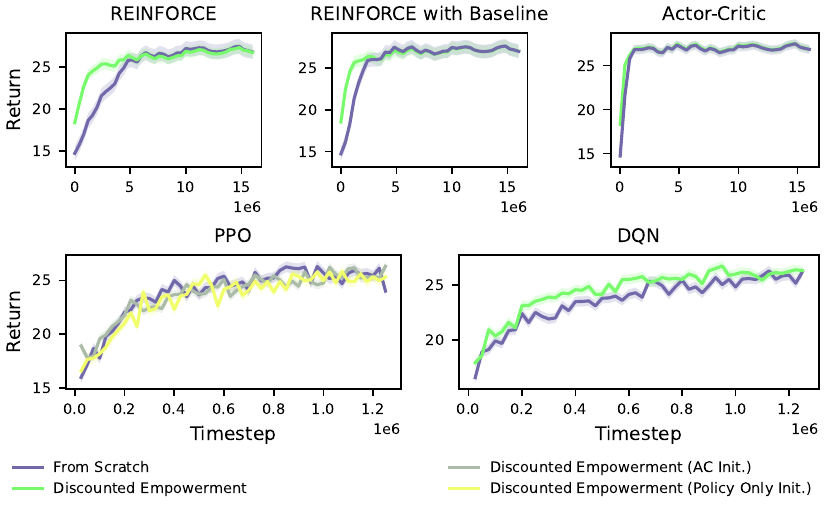}
	\end{subfigure}
	\caption{\label{fig:exp_gridworld} The gridworld results demonstrate that empowerment-based agents consistently outperform agents trained from scratch, with the most significant improvement observed in the case of REINFORCE and Actor-Critic. In contrast, the observed differences for PPO and DQN are comparatively minor. We presume that while empowerment contributes to variance reduction during learning, its impact diminishes in algorithms such as PPO and DQN, which already incorporate effective variance reduction mechanisms.}
	\vspace{-2ex}
\end{figure*}

We now focus on discounted empowerment as our pre-training reward and compare the performance of several RL algorithms within the deterministic gridworld environment. The results of the comparison of these methods are presented in Figure~\ref{fig:exp_gridworld}.

\textbf{REINFORCE (Policy only) \cite{williams_simple_1992}.}
The empirical results demonstrate that agents pre-trained with empowerment-based objectives consistently outperform from-scratch trained REINFORCE agents in terms of data efficiency. This highlights the utility of empowerment as a pre-training signal.

\textbf{REINFORCE with Baseline \cite{williams_simple_1992}.}
Additionally, we train REINFORCE agents with a baseline which can reduce the variance of the policy gradient estimates. Previous work on fine-tuning report results that naive policy initialization with critics often result in suboptimal performance due to catastrophic forgetting \cite{uchendu_jump-start_2023}.
Our results show that while introducing a baseline accelerates convergence, empowerment-based pre-training still provides additional benefits. In particular, the data efficiency improvements from empowerment remain consistent, matching or exceeding those observed with REINFORCE agents without a baseline.

\textbf{On-Policy Actor-Critic \cite{sutton_reinforcement_2018}.} 
While empowerment-based pre-training continues to perform favorably, the performance gap compared to actor-critic agents trained from scratch becomes smaller. This is expected, as our actor-critic method incorporates TD learning for updating both policy and value functions, which introduces additional variance reduction beyond what a baseline alone provides. As a result, the relative advantage of empowerment diminishes somewhat, though it remains a beneficial addition.

\textbf{PPO (On-Policy Actor-Critic) \cite{schulman_proximal_2017}.}
The results obtained with PPO agents indicate that policies fine-tuned using this advanced actor-critic framework exhibit distinct behavioral patterns compared to those fine-tuned with REINFORCE. Notably, we observe that empowerment-based pre-training does not lead to a clear performance improvement in PPO. However, it also does not degrade performance, indicating that empowerment remains a valid and safe pre-training objective. 
Furthermore, we observe comparable fine-tuning performance when only the policy is initialized with pre-trained weights and the critic is re-initialized with random weights, suggesting that PPO's performance is robust to the critic's initialization.

\textbf{DQN (Value-based Off-Policy RL) \cite{mnih_human-level_2015}.}
The experiments involving DQN agents use a value-based, off-policy reinforcement learning approach, where the policy is implicitly derived from the state-action value function. Our results show that empowerment-based pre-training not only accelerates learning but also leads to improved final performance. This indicates that, beyond enhancing data efficiency, empowerment can guide the agent toward better solutions in the DQN setting.

From these results, we assume that empowerment can have an effect similar to a variance reduction mechanism by stabilizing the learning process. While methods like PPO and DQN already include strong built-in variance reduction techniques—such as TD learning~\cite{sutton_reinforcement_2018} and generalized advantage estimation~\cite{schulman_high-dimensional_2016}—empowerment consistently supports or enhances performance. Rather than being redundant, it provides a promising, complementary mechanism to further address the high-variance challenges in reinforcement learning. We suppose that this effect is especially valuable in complex environments where standard methods may struggle early in training, highlighting empowerment as a robust and general-purpose enhancement across algorithms.

\begin{wrapfigure}{r}{0.382\linewidth}
	\includegraphics[width=\linewidth]{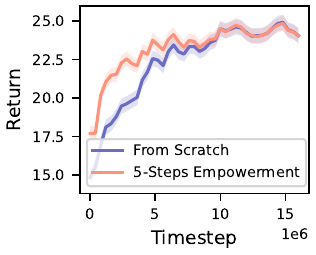}
    \caption{\label{fig:exp_stochastic} REINFORCE training curves in a stochastic gridworld, demonstrating the effectiveness of empowerment-based pretraining also for complex stochastic environments.}
\end{wrapfigure}
\subsection{Stochastic Gridworld}
We now assess the effectiveness of empowerment-based pre-training in a stochastic gridworld environment. In comparison to the deterministic experiments presented above, the stochastic setup provides a more challenging testbed for evaluating the robustness of pre-trained policies and aims to determine whether empowerment-based pre-training enhances the agent's ability to adapt and perform effectively under stochastic dynamics.
Specifically, we compare the fine-tuning performance of REINFORCE agents pre-trained with an empowerment horizon of $5$ steps against agents trained from scratch without any pre-training. The results, presented in Figure~\ref{fig:exp_stochastic}, indicate that agents pre-trained with a $5$-steps empowerment horizon exhibit superior performance in terms of data efficiency compared to those trained from scratch.
Our results demonstrate the effectiveness of empowerment in deterministic as well as in stochastic environments, underscoring its potential as a pre-training signal for enhancing the performance of RL agents.

\section{Discussion} \label{sec:discussion}
This paper explores the concept of empowerment as a foundational principle for pre-training RL agents. By extending the traditional notion of empowerment to include our discounted formulation, we provide a more nuanced understanding of how agents can exert control over their environments across varying temporal horizons without any external knowledge. 
Our experiments demonstrate the efficacy of empowerment-based pre-training in both deterministic and stochastic settings, highlighting its potential to improve data efficiency and adaptability in downstream tasks. In particular, our discounted formulation of multi-step empowerment proves to be an effective pre-training reward, offering strong performance while mitigating the need for hyperparameter tuning.
Additionally, our analysis of empowerment-based pre-training across diverse RL algorithms—including REINFORCE, AC, PPO, and DQN—demonstrates that empowerment consistently performs on par with or better than standard training, often accelerating convergence and, in some cases, leading to improved final performance. These benefits are consistent with the interpretation of empowerment as a form of variance reduction, offering a robust and complementary strategy across both policy- and value-based methods. Another key contribution of this work is the distinction between capacity-achieving and capacity-maximizing policies.
While capacity-maximizing policies exhibit greater data efficiency, both approaches outperform baseline methods, validating the general utility of empowerment as a pre-training signal.

In conclusion, this study establishes empowerment as a robust and versatile pre-training objective, capable of accelerating learning and improving performance across a range of RL algorithms. Our results show that empowerment consistently performs on par with or better than standard training, and its variance-reducing properties make it a compelling addition to existing learning pipelines. While gains with advanced actor-critic methods are sometimes limited, we believe this reflects the strength of these baselines rather than a limitation of empowerment itself. In more complex, high-dimensional environments—where variance is harder to control and exploration more challenging—we expect the benefits of empowerment-based pre-training to become even more pronounced. Realizing this potential will require tackling the computational challenges associated with estimating empowerment efficiently at scale, which we leave for future work. Overall, our findings position empowerment as a promising direction for further research, particularly in the broader context of unsupervised pre-training for large-scale RL agents.

\textbf{Broader impact.} As a general pre-training method for RL, this approach is not inherently tied to any specific application; however, we acknowledge that fine-tuned instances of such pre-trained policies may still pose risks and produce harmful outcomes depending on their deployment context.

In value-based methods, the value learned during pre-training is not the same as the value function for the downstream task. However, this doesn’t seem to affect the performance, can you comment on that, why should this work for value-network initialization?
\begin{ack}
\label{sec:acknowledgements}
Joschka Boedecker is part of BrainLinks-BrainTools which is funded by the Federal Ministry of Economics, Science and Arts of Baden-Württemberg within the sustainability program for projects of the excellence initiative II.
\end{ack}
\newpage
\clearpage

\bibliography{references}
\bibliographystyle{unsrtnat}

\newpage
\clearpage
\appendix
\clearpage
\section{Implementation Details}\label{sec:appx_implementation_details}
\subsection{Stochastic Gridworld}
Stochasticity is introduced by allowing the environment, with a probability of $20$\%, to move the agent randomly either to the left or right relative to its intended direction (i.e. with a probability of $10$\% to the left and $10$\% to the right).
\begin{figure}[!h]
	\centering
	\begin{subfigure}[t]{0.4\linewidth}
		\includegraphics[width=\linewidth]{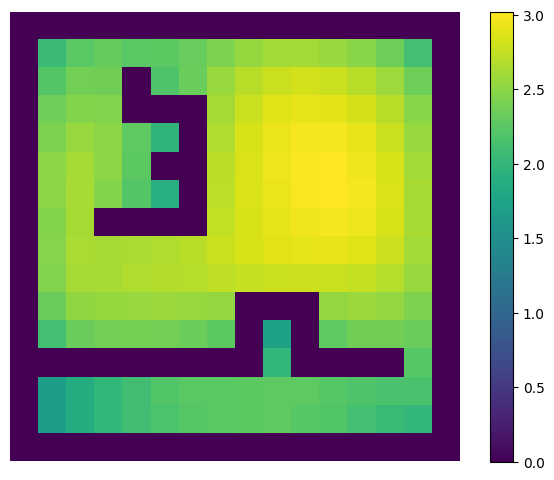}
	\end{subfigure}
	\hspace*{10pt}
	\begin{subfigure}[t]{0.4\linewidth}
		\includegraphics[width=\linewidth]{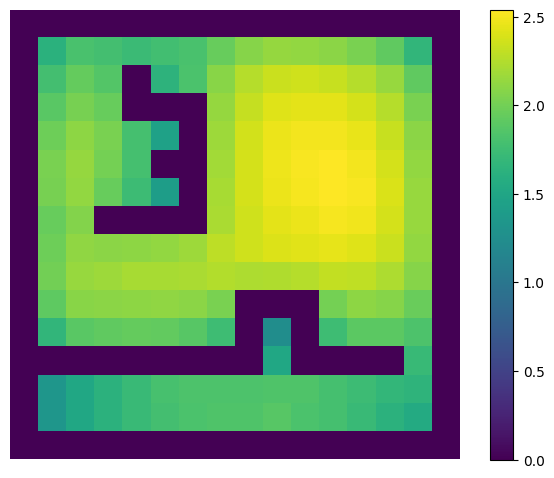}
	\end{subfigure}
	\label{fig:gridworld_stochastic}
	\caption{$5$-steps empowerment-values of the stochastic gridworld environment. In the left environment the agent randomly moves to the left or right with a probability of $10$\% and in the right environment the agent randomly moves to the left or right with a probability of $20$\%. Otherwise the agent moves in the intended direction of the chosen action.}
\end{figure}

\subsection{Experiments}\label{sec:appx_experiments}
Each episode in our gridworld environment consists of $32$ steps. Rewards are normalized to the range $[0,1]$, with the goal state assigned a reward of $1$. In empowerment-maximization environments, the goal state corresponds to the state with maximal empowerment. We calculate discounted empowerment with a discount factor of $\lambda=0.95$ and a time horizon of $H=32$ steps. To ensure exploration during later fine-tuning, we penalize low entropy of the policy during pre-training and fine-tuning. All our networks are trained using the Adam optimizer \cite{kingma_adam_2017} with default parameters.

We evaluate each agent every $25,000$ environment steps for $5$ (REINFORCE and AC) or $25$ (PPO and DQN) episodes. The graphs in our plots show the mean return and the standard deviation of the mean over the evaluation results of all runs of the experiment. 

Our implementation is based on JAX \cite{bradbury_jax_2018}. PPO and DQN agents are trained using NVIDIA V100 GPUs, with each training run requiring approximately two hours using a single GPU. We utilize the Stable-Baselines (SBX) framework \cite{raffin_stable-baselines3_2021} for these two algorithms. All other algorithms are executed on CPUs, with each run completing in approximately $45$ minutes (for pre-training and fine-tuning). Across all experiments, we allocate $80$ GB of memory; however, lower memory usage may suffice depending on the specific algorithm.

\subsubsection{Training of the Capacity-Achieving Policy}
The capacity-achieving policy is the output of the Blahut-Arimoto algorithm \cite{blahut_computation_1972,arimoto_algorithm_1972}. The algorithm provides only action distributions for each state independently. We pre-train our neural network policies by approximating these distributions through behavior cloning from sampled actions, enabling effective subsequent fine-tuning.
\clearpage

\subsection{Hyperparameters}
\begin{table}[h]
    \centering
    \caption{\label{tab:hyperparameters} Overview of the hyperparameters for REINFORCE.}
    \begin{tabularx}{0.6\textwidth}{@{}ll@{}}
        \toprule
                                           & \textbf{REINFORCE}    \\
        \midrule
        \textbf{General}                   &                       \\
        Random seeds                       & $5$                   \\
        Number of environments             & $16$                  \\
        Pretraining steps                  & $16 \times 1,000,000$ \\
        Finetuning steps                   & $16 \times 1,000,000$ \\
        Batch size                         & $32$                  \\
        \midrule
        \textbf{Actor}                     &                       \\
        Hidden layers                      & $2$                   \\
        Hidden layer dim.                  & $256$                 \\
        Learning Rate                      & $1 \times 10^{-4}$    \\
        Entropy coefficient                & $1 \times 10^{-1}$    \\
        Discount factor $\gamma$           & $0.99$                \\
        \midrule
        \textbf{Baseline (if used)} &                       \\
        Hidden layers                      & $2$                   \\
        Hidden layer dim.                  & $256$                 \\
        Learning rate                      & $3 \times 10^{-4}$    \\
        \bottomrule
    \end{tabularx}
\end{table}
\begin{table}[h]
    \centering
    \caption{\label{tab:hyperparameters} Overview of the hyperparameters for AC.}
    \begin{tabularx}{0.6\textwidth}{@{}ll@{}}
        \toprule
                                           & \textbf{AC}    \\
        \midrule
        \textbf{General}                   &                       \\
        Random seeds                       & $5$                   \\
        Number of environments             & $16$                  \\
        Pretraining steps                  & $16 \times 1,000,000$ \\
        Finetuning steps                   & $16 \times 1,000,000$ \\
        Batch size                         & $32$                  \\
        \midrule
        \textbf{Actor}                     &                       \\
        Hidden layers                      & $2$                   \\
        Hidden layer dim.                  & $256$                 \\
        Learning Rate                      & $1 \times 10^{-4}$    \\
        Entropy coefficient                & $1 \times 10^{-1}$    \\
        Discount factor $\gamma$           & $0.99$                \\
        \midrule
        \textbf{Critic} &                       \\
        Hidden layers                      & $2$                   \\
        Hidden layer dim.                  & $256$                 \\
        Learning rate                      & $1 \times 10^{-4}$    \\
        \bottomrule
    \end{tabularx}
\end{table}
\begin{table}[h]
    \centering
    \caption{\label{tab:hyperparameters_ppo} Overview of the hyperparameters for PPO.}
    \begin{tabularx}{0.6\textwidth}{@{}ll@{}}
        \toprule
                                    & \textbf{PPO}                             \\
        \midrule
        \textbf{General}            &                                          \\
        Random seeds                & $6$                                      \\
        Number of environments      & $4$                                      \\
        Pretraining steps           & $1,000,000$                              \\
        Finetuning steps            & $1,250,000$                              \\
        Batch size                  & $64$                                     \\
        GAE-$\lambda$               & $0.95$                                   \\
        Advantage normalization     & \texttt{True}                            \\
        Number of epochs per update & $10$                                     \\
        Steps between updates       & $1024$                                   \\
        Max. gradient norm clipping & $0.5$                                    \\
        \midrule
        \textbf{Actor}              &                                          \\
        Clip range                  & $0.2$                                    \\
        CNN architecture            & Same as in \citet{mnih_human-level_2015} \\
        Learning rate               & $1 \times 10^{-4}$                       \\
        Entropy coefficient         & $1 \times 10^{-2}$                       \\
        Discount factor $\gamma$    & $0.99$                                   \\
        \midrule
        \textbf{Critic}             &                                          \\
        Critic loss weight coeff.   & $0.5$                                    \\
        CNN architecture            & Same as in \citet{mnih_human-level_2015} \\
        Learning rate               & $1 \times 10^{-4}$                       \\
        \bottomrule
    \end{tabularx}
\end{table}

\begin{table}[h]
    \centering
    \caption{\label{tab:hyperparameters_dqn} Overview of the hyperparameters for DQN.}
    \begin{tabularx}{0.6\textwidth}{@{}ll@{}}
        \toprule
                                     & \textbf{DQN}                             \\
        \midrule
        \textbf{General}             &                                          \\
        Random seeds                 & $6$                                      \\
        Number of environments       & $1$                                      \\
        Replay buffer size           & $1,000,000$                              \\
        Pretraining steps            & $1,000,000$                              \\
        Finetuning steps             & $1,250,000$                              \\
        Batch size                   & $32$                                     \\
        Number of epochs per update  & $10$                                     \\
        Steps between updates        & $4$                                      \\
        Seed steps                   & $100$                                    \\
        Initial exploration rate     & $1.0$                                    \\
        Final exploration rate       & $0.01$                                   \\
        Decay exploration rate until & $100,000$ steps                          \\
        \midrule
        \textbf{Critic}              &                                          \\
        Discount factor $\gamma$     & $0.99$                                   \\
        Target critic $\tau$         & $0.95$                                   \\
        Target update interval       & Every $1,000$ steps                      \\
        CNN architecture             & Same as in \citet{mnih_human-level_2015} \\
        Learning rate                & $1 \times 10^{-4}$                       \\
        \bottomrule
    \end{tabularx}
\end{table}

\clearpage

\section{Information Geometry of Empowerment-based RL}
\label{sec:appx_info_geom_empowerment}
\begin{figure*}[h]
	\centering
	\begin{subfigure}[t]{0.45\linewidth}
		\includegraphics[width=\linewidth]{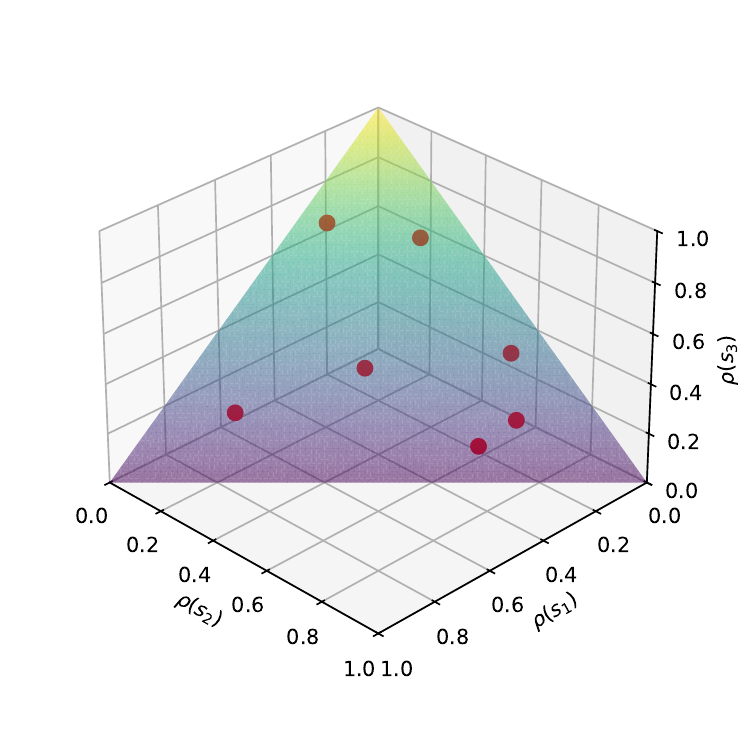}
	\end{subfigure}
	\begin{subfigure}[t]{0.45\linewidth}
		\includegraphics[width=0.95\linewidth]{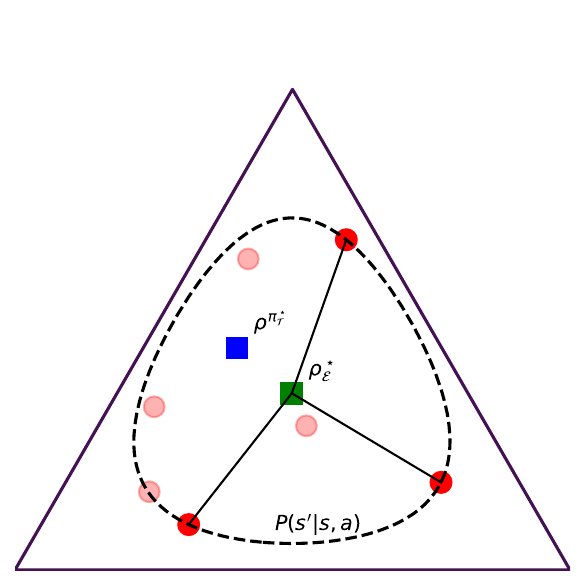}
	\end{subfigure}
	\caption{\label{fig:info_geom} The left panel illustrates the state-marginal simplex of an environment comprising three discrete states, while the right panel depicts the corresponding state-marginal polytope. The green point represents the initialization of an agent that has been pre-trained using empowerment objectives. All other points located within the information ball correspond to feasible state-marginals, encompassing all possible initializations as well as all achievable task-specific policies. The dark red points indicate policies positioned along the boundary of the information ball, representing those that are maximally distant from the empowerment-pretrained agent. Notably, because the empowerment-pretrained agent is situated near the center of the information ball, it is expected to exhibit a high degree of adaptability, enabling it to transfer and adapt to a wide range of downstream tasks more rapidly on average. Adapted from \cite{eysenbach_information_2022}.}
\end{figure*}

We now want to give an intuition why it is beneficial from an information geometric point of view to maximize mutual information. We build on prior work to provide an information geometric interpretation of what mutual information maximization does \cite{eysenbach_information_2022}. But in contrast we transfer the interpretation from latent skills to ordinary actions.

The average state distribution can be written as
$$\rho(\mathbf{s}' | \mathbf{s}) = \sum_\mathbf{a} \rho(\mathbf{s}', \mathbf{a} | \mathbf{s}) = \sum_\mathbf{a} \mathcal{P}(\mathbf{s}' | \mathbf{s}, \mathbf{a}) \pi(\mathbf{a} | \mathbf{s}).$$

\begin{remark}
    \textup{(Theorem 13.1.1 from \citet[page 430]{cover_elements_2005})}
    Maximizing mutual information is equivalent to minimizing the divergence between the average state distribution $\rho(\mathbf{s'} \mid \mathbf{s})$ and the furthest achievable state distributions \cite{eysenbach_information_2022,cover_elements_2005}:
    $$\mathcal{E}(s)=\max_{\pi(\mathbf{a} | \mathbf{s})} I(\mathbf{s}';\mathbf{a} | \mathbf{s}) = \min_{\rho(\mathbf{s}' | \mathbf{s})} \max_a D_{KL} (\rho(\mathbf{s}'| \mathbf{s}, \mathbf{a}) | \rho(\mathbf{s}' | \mathbf{s}))$$
\end{remark}

In other words: By maximizing mutual information, we minimize the maximum distance of the average state distribution $\rho(\mathbf{s}' | \mathbf{s})$ to all other achievable state distributions (under the information ball given by $\mathcal{P}(\mathbf{s}'|\mathbf{s}, \mathbf{a})$) \cite[page 430, Theorem 13.1.1]{cover_elements_2005}.

Thus, the optimal prior of $\rho(\mathbf{s}' | \mathbf{s})$ that achieves the minimum is the output distribution $\rho^*(\mathbf{s}' | \mathbf{s})$ induced by the capacity-achieving input distribution $\pi^*(\mathbf{a} | \mathbf{s})$:
$$\rho^*(\mathbf{s}' | \mathbf{s}) = \sum_\mathbf{a} \mathcal{P}(\mathbf{s}' | \mathbf{s}, \mathbf{a}) \pi^*(\mathbf{a} | \mathbf{s}) = \mathbb{E}_{\pi^*(\mathbf{a} | \mathbf{s})} [\mathcal{P}(\mathbf{s}' | \mathbf{s}, \mathbf{a})]$$
Thus, the empowerment-achieving policy has a state marginal distribution in the center of the information ball bringing the average state distribution close to other policies.
\citet{eysenbach_information_2022} argues that, in context of skill learning, such an average state distribution would be a beneficial initialization for downstream task learning. Whether such an initialization is suitable for unconditioned policy pre-training—i.e., policies not conditioned on skills—remains an open question, as a fundamental distinction exists between actions, which are explicitly grounded in the MDP, and skills, which are abstract latent variables.
\clearpage
\end{document}